%% file: egbib.tex
\documentclass[10pt,twocolumn,letterpaper]{article}

\usepackage{style/cvpr}
\usepackage{times}
\usepackage{epsfig}
\usepackage{graphicx}
\usepackage{amsmath}
\usepackage{amsfonts}
\usepackage{bbm}
\usepackage{amssymb}
\usepackage{adjustbox}
\usepackage{subcaption}
\usepackage{ctable}
\usepackage{multicol,boldline}
\usepackage{makecell}
\usepackage{lipsum}

\input{utils/macros.tex}

\input{utils/symbols.tex}
\newcommand\blfootnote[1]{%
  \begingroup
  \renewcommand\thefootnote{}\footnote{#1}%
  \addtocounter{footnote}{-1}%
  \endgroup
}

\usepackage[pagebackref=true,breaklinks=true,letterpaper=true,colorlinks,bookmarks=false]{hyperref}

\cvprfinalcopy 

\def\cvprPaperID{5785} 

\ifcvprfinal\fi
\begin{document}

\title{Polygonal Point Set Tracking}

\author{Gunhee Nam$^{1*}$
\and    Miran Heo$^{2}$
\and    Seoung Wug Oh$^{3}$
\and    Joon-Young Lee$^{3}$
\and    Seon Joo Kim$^{2}$}


\maketitle
   
\blfootnote{$^*$Work mostly done during a M.S. student at Yonsei University}
\input{body/01_abstract}
\vspace{-0.3cm}
\input{body/02_introduction}
\input{body/03_related_work}
\input{body/04_method}
\input{body/05_experiments}
\input{body/06_conclusion}

\input{body/07_acknowledgements}
\input{supple/supple_main}

{\small
\bibliographystyle{utils/ieee_fullname}
\bibliography{egbib}
}

\end{document}

%% file: utils/macros.tex
\newcommand{\Tref}[1]{Table~\ref{#1}}

\newcommand{\Fref}[1]{Figure~\ref{#1}}

\newcommand{\Sref}[1]{Section~\ref{#1}}

%% file: body/01_abstract.tex
\begin{abstract}


In this paper, we propose a novel learning-based polygonal point set tracking method.
Compared to existing video object segmentation~(VOS) methods that propagate pixel-wise object mask information, we propagate a polygonal point set over frames. 
Specifically, the set is defined as a subset of points in the target contour, and our goal is to track corresponding points on the target contour.
Those outputs enable us to apply various visual effects such as motion tracking, part deformation, and texture mapping.
To this end, we propose a new method to track the corresponding points between frames by the global-local alignment with delicately designed losses and regularization terms.
We also introduce a novel learning strategy using synthetic and VOS datasets that makes it possible to tackle the problem without developing the point correspondence dataset.
Since the existing datasets are not suitable to validate our method, we build a new polygonal point set tracking dataset and demonstrate the superior performance of our method over the baselines and existing contour-based VOS methods.
In addition, we present visual-effects applications of our method on part distortion and text mapping.

\end{abstract}

%% file: body/02_introduction.tex
\section{Introduction}
\label{introduction}

Object mask tracking in a video is one of the most frequently required tasks in visual effects~(VFX). 
However, the task (\ie rotoscoping) is so painstaking and time-consuming that even a highly-skilled designer processes only a dozen frames on average per day \cite{li2016roto++}.
Therefore, propagating object mask information through subsequent frames becomes a critical problem to reduce human labor for rotoscoping.
Propagation methods are categorized into four groups based on object representations: point, region, contour, and polygonal point set~(\Fref{fig:comp}). Each representation carries different amount of information.

\input{body/figures/teaser}

\input{body/figures/task_comparison}

Patch tracking~\cite{danelljan2014kcf, lukezic2017csrt, danelljan2016dcf, bolme2010mosse} denotes a target object as \textit{point} representation and tracks a target point over frames by matching the patch around the point.
The tracking enables visual effects that require positional information such as motion tracking and texture mapping.
These applications often require multiple patch tracking to compute point-to-point information.
However, conducting each patch tracking independently ignores strong correlations between target points, thus multiple patch tracking are susceptible to a drift problem and are not suitable for mask propagation.

Meanwhile, video object segmentation~(VOS)~\cite{voigtlaender2017online-vos, caelles2017oneshot-vos, jampani2017biliteral-vos, wug2018fast-vos, oh2019stm, miao2020manet} and contour tracking~\cite{isard1996contourtracking, yilmaz2004levelset, chen2006multicuecontour, saboo2020deeprotoscope} propagate target object information over subsequent frames by representing the target as a \textit{region}~(\ie mask) and a \textit{contour} respectively.
These representations can describe only the target area without any pixel correspondences, therefore they are not applicable to complex VFX scenarios that require point-to-point relation information (\eg, \Fref{fig:teaser}).

On the other hand, \textit{polygonal point set} tracking combines the positional information with the target region.
The polygonal point set is defined as a subset of contour points that represent an object in a polygonal shape. 
By tracking the point set, we can get both the object contour and the point-to-point matching information. 
Previous works in this category~\cite{agarwala2004keyframe, li2016roto++, lu2016cpc, perez2019roam} focus on making the user interaction easier for highly customizable results rather than taking the point-to-point matching (or tracking) into account.
Therefore, they assume heuristic shape priors of an object and often exhibit propagation failures for challenging object motions.

In this paper, we aim to track all points directly through a learning-based approach without assuming a heuristic shape prior and propose a novel point set tracking method.
%
According to the hypothesis that the target object state in adjacent frames is highly correlated, we train a network to learn progressive alignment of a point set between frames. 
We first match the point set globally using a simple rigid transformation.
Then, we further tune each point position in a coarse to fine manner using a local alignment module.
Our local alignment module~(LAM) adopts recurrent neural networks~(RNN) to take into account the temporal history of each point and also uses multi-head attention~(MHA) modules for non-local communication among the points.
In addition, we regularize the alignments to avoid drifting and honor the original topology of a target polygon in challenging situations.

We introduce a new learning strategy to train our model without fully annotated data. 
Currently available VOS datasets~\cite{caelles2018davis, xu2018youtube-vos} only contain region information (\ie masks), thus it cannot be directly applied to our point correspondence learning.
To overcome the data issue, we propose an unsupervised learning method based on cyclic consistency between predicted frames~\cite{wang2019unsupervisedtracking, wang2019correspondence}.
We also obtain the supervision for point set tracking by synthesizing data from image instance segmentation data.
To the best of our knowledge, this is the first work on learning-based point set tracking that considers the point correspondence.

\Fref{fig:teaser} shows an example application that utilizes our network results.
In this example, to exaggerate the upper body, the same effect is applied to the target across the entire frame, even if the user edits the points in the set individually only in the first frame.

Popular evaluation datasets for video object segmentation are not suitable for evaluating point-set tracking as they only provide mask annotations~\cite{caelles2018davis, fan2015jumpcut, li2013segtrack_v2, lu2016cpc}.
While CPC~\cite{lu2016cpc} provides annotations of object contours, it is also not sufficient for the evaluation as there is no point correspondence annotation. 
To this end, we introduce an evaluation dataset for polygonal point set tracking, consisting of 30 sequences.
To build the dataset, named PoST, we augment video clips from existing VOS datasets with additional point set annotations with correspondence.
We evaluate our method on PoST and the existing VOS datasets~\cite{caelles2018davis, lu2016cpc}, and we show that our method outperforms competing methods with a large margin.

Our contributions are summarized as follows:
\begin{itemize}
\item We propose a novel learning-based method for polygonal point set tracking with point correspondence for the first time.
\item We design a local alignment module for point tracking with taking temporal history and communication with other points into account.
\item We present a learning strategy to train a deep network for point set tracking using unsupervised learning and synthesized data.
\item We introduce a new dataset for evaluating performance of point set tracking.
\end{itemize}

%% file: body/figures/teaser.tex
\begin{figure}[t]
\begin{center}

\includegraphics[width=1.0\linewidth]{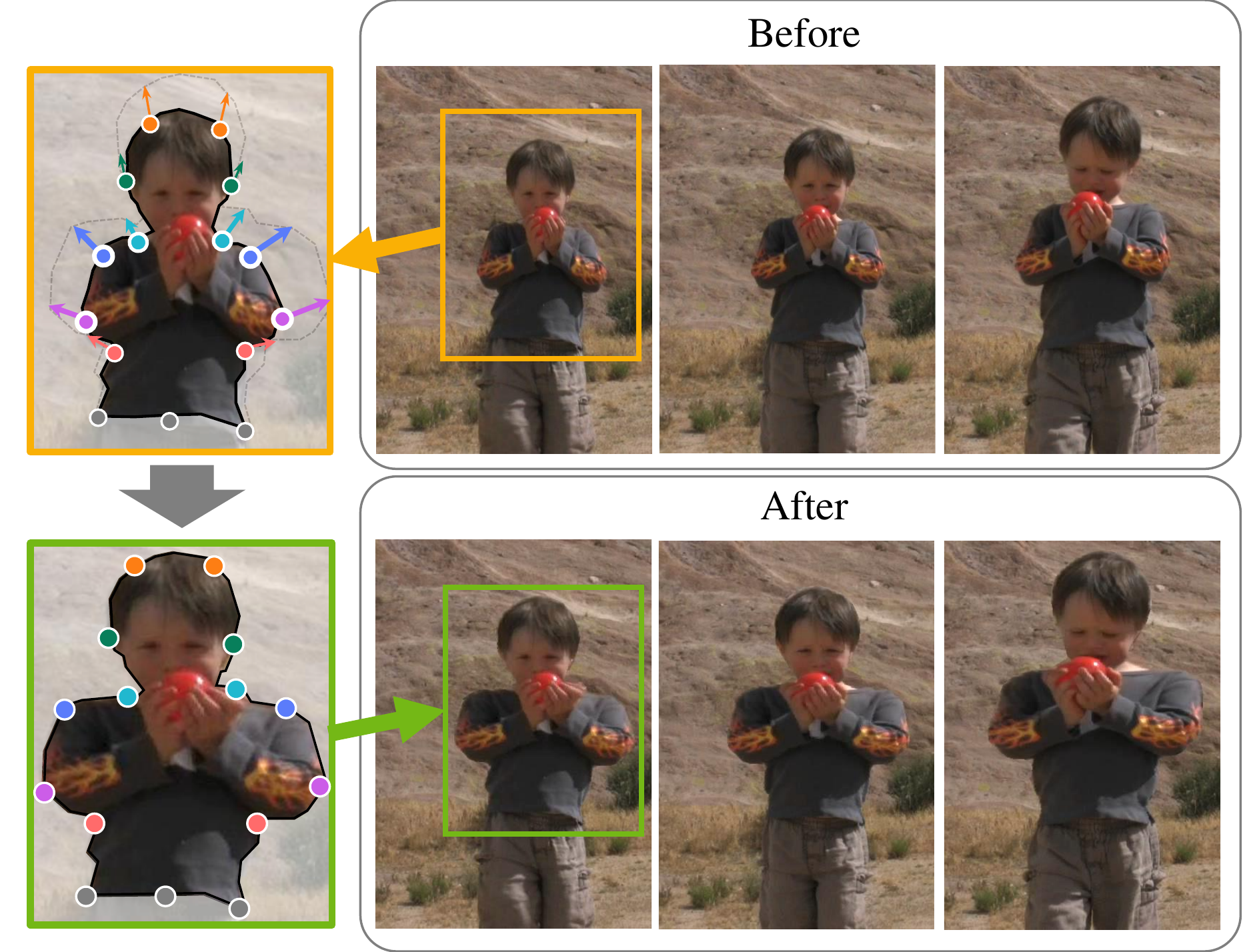}
\end{center}
\vspace{-0.5cm}
   \caption{Our method tracks a set of points in a polygon over frames. The output represents mask contour with point correspondences across frames. It allows multiple applications, \eg, a non-rigid transformation of a specific part of an object over time as shown here.}
\label{fig:teaser}
\vspace{-0.5cm}
\end{figure}

%% file: body/figures/task_comparison.tex
\begin{figure*}[t]
\begin{center}

\includegraphics[width=.95\linewidth]{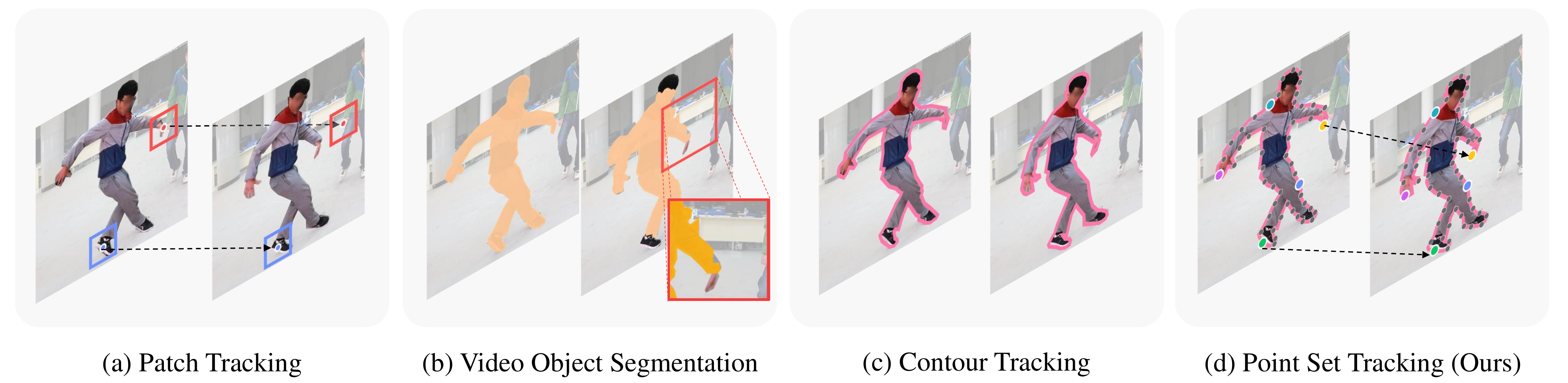}
\end{center}
\vspace*{-0.6cm}
   \caption{Illustration of different approaches for object mask propagation. (a)~Independent multi-patch tracking represents an object coarsely and drifts easily. (b)~ Region-based video object segmentation achieves high accuracy in pixel-level dense prediction, but it usually may yield cattery boundaries due to a high degree of freedom and does not provide point correspondences. (c)~In contour tracking, the constrained contour representation can give us clean boundary, but point correspondence information is still missing. On the other hand, (d)~Polygonal point set tracking provides both clean polygonal object mask and point correspondences across frames.}
\label{fig:comp}
\vspace{-0.5cm}
\end{figure*}

%% file: body/03_related_work.tex
\section{Related Work}


\noindent\textbf{Object Mask Propagation}
Patch tracking is a naive approach of the object mask propagation at the point level.
By tracking a given patch through all frames, the patch location provides the positional information of the tracked part of an object across frames.
However, recent research based on a deep learning approach has focused on object-level tracking rather than patch-level because it is hard to annotate corresponding patches over frames ~\cite{bertinetto2016siamfc, bhat2019dimp, danelljan2020prdimp}.
Although optimization-based methods~\cite{danelljan2014kcf, lukezic2017csrt, danelljan2016dcf, bolme2010mosse} bypass this data issue, it is not robust enough since they rely on hand-crafted features. 
Moreover, while multiple patch tracking is performed on the different parts of the same object in the case of VFX, such as texture mapping, they drift easily because each patch is tracked independently.

On the other hand, region-based video object segmentation estimates pixel-wise masks.
In this approach, the user-supplied mask is temporally propagated to other frames to aid the time-consuming per frame segmentation.
Recently, thanks to the representational power of deep learning, the region-based methods have reached a milestone in its object mask propagation performance~\cite{voigtlaender2017online-vos, caelles2017oneshot-vos, jampani2017biliteral-vos, wug2018fast-vos, oh2019stm, miao2020manet}.
Despite the success of the aforementioned methods, its shape representation inherently limits many editing applications as it cannot provide point-to-point information.

Contour tracking methods~\cite{isard1996contourtracking, yilmaz2004levelset, chen2006multicuecontour, saboo2020deeprotoscope} also propagate the object masks but use a more constrained representation~(\ie object boundary).
Contour tracking has been performed by probabilistic~\cite{isard1996contourtracking, yilmaz2004levelset} and hidden Markov models~\cite{chen2006multicuecontour}.
More recently, Saboo~\etal~\cite{saboo2020deeprotoscope} propose a learning-based framework that solves the problem by adopting an attention mechanism. 
By employing the contour representation, these methods reduce prediction noise and yield a clean object boundary but still lack temporal point-to-point correspondence similar to the region-based object segmentation methods.
 
In polygonal point set tracking, different from region-based segmentation and contour tracking, each point can be tracked. 
In this problem setting, various cues for propagating an object mask are previously explored. 
In \cite{agarwala2004keyframe}, an interpolation between two key frames is performed. 
Some methods try to find sharp object edges using snakes~\cite{blake2012active_contours, kass1988snakes} after a global tracking through either using shape manifold~\cite{li2016roto++} or shape prior~\cite{lu2016cpc, perez2019roam}.
However, the previous point set tracking methods rather focus on its convenience for user interaction for controlling object shapes than establishing an accurate point-to-point correspondence.  
Thus, many of them are limited to applications that require matching points over time.

\noindent\textbf{Point Set Representation in Deep Learning}
While point set representation is not popular in modern deep learning architectures in computer vision, there are some efforts to employ it.
To find an efficient way against manual annotation for segmentation, the shape representation is defined as polygon structure in \cite{castrejon2017polygon-rnn, acuna2018polygon-rnn++, ling2019curve-gcn}.
In these methods, several architectures of recurrent neural networks~(RNN)~\cite{castrejon2017polygon-rnn}, graph neural network~(GNN)~\cite{acuna2018polygon-rnn++} and graph convolution network~(GCN)~\cite{ling2019curve-gcn} are suggested to deal with point set in deep approach.
Point set representation is also employed for image instance segmentation as well~\cite{peng2020deep_snake, liang2020polytransform}.
Other options to handle the representation are proposed in those methods such as circular convolution~\cite{peng2020deep_snake} and transformer~\cite{liang2020polytransform}.
These previous methods show considerable potential for the shape representation of point set but only focus on a single image level.
Point set tracking is inherently impossible for these image-level approaches.

\noindent\textbf{Global-local alignment}
Since adjacent frames in a video sequence are highly correlated, global-local alignment is a common approach in many video-related tasks such as optical flow~\cite{sun2018pwc}, video inpainting~\cite{lee2019copy-and-paste}, and object tracking~\cite{fan2019siamcascade}.
Although not a learning-based nor point-to-point tracking method, SnapCut~\cite{bai2009snapcut} is also one of the most popular rotoscoping approaches that shares a similar philosophy with ours, where patches are tracked first and then the local classifier refines the contour.
Following this concept, we adopt the global-local alignment in our framework.


%% file: body/04_method.tex
\input{body/figures/overview}


\section{Method}
Our method tracks a set of polygonal points representing a subset of points on the contour of the target object.
As shown in \Fref{fig:overview}, our framework is divided into two steps: global and local alignments.
For global alignment, we compute an affine transformation matrix between the previous and the current frames.
We globally align the previous frame and its point set using the computed matrix.
In the second step, we forward the transformed frame and point set to a local feature encoder, and extract point-wise features for each point in the set. We forward the extracted point features to Local Alignment Module (LAM) and update the point correspondences locally.
LAM computes the offset of each point to update the point's location progressively in a coarse to fine manner. 

\subsection{Global Alignment}
Since the deformation of an object or viewpoint change is small between adjacent frames, a simple geometric transformation like affine transform can align the two frames to some extent. 
Inspired by \cite{jaderberg2015spatial_transformer, lee2019copy-and-paste}, the global alignment network predicts an affine transformation matrix to align the previous frame $\mathcal{I}_{t-1}$ toward the current frame $\mathcal{I}_t$.
To estimate the affine transformation matrix $\mathbf{A}_{t-1 \rightarrow t}$, the binary target mask $\hat{\mathcal{M}}_{t-1}$ obtained from the polygonal point set $\hat{\mathbf{P}}_{t-1}$ is also given as an additional input into the global alignment network, allowing it to focus on the target while disregarding the background.
For computational efficiency, the input is resized into a half and a lightweight backbone is used for the global alignment network as in \cite{lee2019copy-and-paste}.
In short, $\mathbf{A}_{t-1 \rightarrow t}$ is obtained by the global alignment network $f_{\text{glob}}(\cdot)$ as follows:
\vspace{-0.3cm}
\begin{equation}
\label{eq:global_alignment}
    \mathbf{A}_{t-1 \rightarrow t} = f_{\text{glob}}(\mathcal{I}_t^\downarrow, \mathcal{I}_{t-1}^\downarrow, \hat{\mathcal{M}}_{t-1}^\downarrow),
\end{equation}
where $\mathcal{I}^\downarrow$ and $\hat{\mathcal{M}}^\downarrow$ denote downsized image and mask respectively. 
The output of the network is a 6-dimensional vector that represents an affine matrix.
The previous frame $\mathcal{I}_{t-1}$ and its point set $\hat{\mathbf{P}}_{t-1}$ are then warped by the computed $\mathbf{A}_{t-1 \rightarrow t}$ into $\mathcal{I}_{t-1 \rightarrow t}$ and $\hat{\mathbf{P}}_{t-1 \rightarrow t}$.

\subsection{Local Alignment}
After the global alignment, each point in the point set is further aligned locally.
We extract point features using a local feature encoder.
Specifically, We use \texttt{resnet50}~\cite{he2016resnet} as an encoder backbone and take FPN feature maps~\cite{lin2017fpn} for accurate localization.
We feed the current frame $\mathcal{I}_t$, the warped previous frame~$\mathcal{I}_{t-1 \rightarrow t}$ and the warped target mask~$\hat{\mathcal{M}}_{t-1 \rightarrow t}$ to the encoder after concatenating them along the channel axis.

From the encoded feature maps, we sample point feature vectors according to the location of each point in the point set.
Given the point vectors, our \textit{Local Alignment Module}~(LAM) calculates the displacement offset of each point.
The locations of the points are updated using the offsets, and we repeat the point feature sampling and the offset update.  
In each iteration, we use the feature maps at a different scale from coarse to the finest levels, thus the point set is aligned in a coarse-to-fine manner. 

\noindent\textbf{Local Alignment Module~(LAM). } 
Local Alignment Module~(LAM) takes a set of point feature vectors and yields 2-channeled offsets (\ie horizontal and vertical), where each 2-channel offset corresponds to each point.
\Fref{fig:overview} describes the detailed architecture of LAM.
The module can be divided into four parts: \textit{positional encoding}, \textit{backbone}, \textit{temporal information transfer}, and \textit{prediction}.

\textit{Positional encoding} allows the network to identify the order of each point in the set.
We use the sinusoidal positional embedding as in \cite{vaswani2017transformer}.
However, in our positional embedding, the first point should meet the last point reflecting the cyclic characteristic of a polygon.
Therefore, we adjust the period of the sinusoidal function according to the number of points $N$ in the point set as follows:
\vspace{-0.1cm}
\begin{equation}
    \mathbf{E}_\text{order}^i = [\text{sin}(2 \pi i / N), \text{cos}(2 \pi i / N)], 
\end{equation}
where $\mathbf{E}_\text{order}^i$ denotes the positional embedding of  $i^\text{th}$ vertex.

\textit{Backbone} is constructed by stacking 8 base blocks where each block consists of two circular convolutions~\cite{peng2020deep_snake} followed by multi-head attention~(MHA)~\cite{vaswani2017transformer}.
In the last layer, features from each level are fused by concatenation and an 1x1 convolution followed by max pooling.
\textit{Temporal information transfer} improves the temporal consistency of the estimation by taking the history of each point.
We use Long Short-Term Memory~(LSTM)~\cite{hochreiter1997lstm} for this purpose.
Finally, \textit{prediction} layer outputs the offset of each point.

\section{Training}
\subsection{Objective Function}
\noindent\textbf{Global Alignment.}
To train the global alignment network, we use two losses:  point set matching loss and pixel matching loss. 
The point set matching loss~\cite{ling2019curve-gcn} considers a group of points by measuring a global distance, where each point of two different point sets is matched one-to-one with each other along the polygonal path.
As there are multiple global distances between two point sets depending on the starting index, we minimize the minimum of all possible distances. 
With a point set with starting index $k$,  $\mathbf{P^k} = [\mathbf{P}^{k\%N}, \mathbf{P}^{(k+1)\%N},...,\mathbf{P}^{(k+N-1)\%N}]$, the point set matching loss $\mathcal{L}_g$ is defined as follows:
\begin{equation}
\small
\label{eq:global_mating_loss}
    \mathcal{L}_g = \min_{k=[0,...,N-1]}\sum_{i=0}^{N-1}{\left\lVert\mathbf{P}^{\mathbf{k}i} - \hat{\mathbf{P}}^{i}\right\lVert_1}, 
\end{equation}
where $\mathbf{P}^i$ and $\hat{\mathbf{P}^i}$ denote the points each from the ground-truth and the globally-aligned point sets with index $i$~($0 \leq i < N$) respectively, and $\left\lVert\cdot\right\lVert$ denotes smooth L1 distance~\cite{girshick2015fast_rcnn}.

Pixel matching loss is based on the hypothesis that the brightness (or color) of the aligned pixel from the previous frame should be similar to that of the corresponding pixel in the current frame ~\cite{lee2019copy-and-paste}.
The pixel matching loss~$\mathcal{L}_{\text{p}}$ is calculated as follows:
\begin{equation}
\small
    \mathcal{L}_{\text{p}} = \frac{1}{K}\sum_{x=0}^{W-1}\sum_{y=0}^{H-1}{\mathbbm{1}^{xy}_\text{obj}\left\lVert \mathcal{I}_t^{xy} - \mathcal{I}_{t-1 \rightarrow t}^{xy}\right\lVert_2},
\end{equation}
where $\mathbbm{1}^{xy}_\text{obj}$ is an indicator function to check if pixel $(x, y)$ belongs to the target mask in the current frame $\mathcal{I}_t$ whose width and height is $W$ and $H$ respectively, and $K = \sum_{x=0}^{W-1}\sum_{y=0}^{H-1}{\mathbbm{1}^{xy}_\text{obj}}$.

\noindent\textbf{Local Alignment.}
We have two different scenarios in training and use different objective functions for the local alignment in each scenario. When we have ground-truth point correspondences (\eg, synthetic data), we use the smooth L1 loss between two matching points as $\mathcal{L}_c = \sum_{i=0}^{N-1}{||\mathbf{P}^i - \hat{\mathbf{P}}^i||_1}$. 

Another scenario is with existing VOS datasets, having only ground-truth masks without point correspondences. In this case, we sample a polygonal point set along the ground-truth mask boundary at each frame. Then, we adopt Chamfer distance between the predicted point set and the target point set as our objective for local alignment. 
The objective encourages each predicted point to be mapped to a point on the object boundary and is formally defined as follows:
\begin{equation}
\small
\begin{split}
    \mathcal{L}_c = \frac{1}{N}\sum_{i=0}^{N-1}&{\min_{j=[0,...,N-1]}{||\mathbf{P}^i - \hat{\mathbf{P}}^j ||_2}} \\ 
    &+ \frac{1}{N}\sum_{j=0}^{N-1}{\min_{i=[0,...,N-1]}{|| \mathbf{P}^i - \hat{\mathbf{P}}^j ||_2}}.
\end{split}
\end{equation}
In our implementation, $M=N$ as we sample the same number of points at each frame.

In addition to the correspondence objective, we include regularization terms into our objective function to avoid drifting in challenging situations, \eg, a corresponding point is occluded. 
We want to honor the previous shape topology in the case, therefore we make use of the first and second derivative regularization~($\mathcal{R}_1$ and $\mathcal{R}_2$) on the predicted point set and prevent dramatic changes in the length and the angle of the point set. 
The regularization terms are defined as follows:
\begin{equation}
\small
    \mathcal{R}_1 = \sum_{i=0}^{N-2}{(|| \hat{\mathbf{P}}_{t}^{i} - \hat{\mathbf{P}}_{t}^{i-1} ||_2 - || \hat{\mathbf{P}}_{t-1}^{i} - \hat{\mathbf{P}}_{t-1}^{i-1} ||_2)^2},
\end{equation}
\begin{equation}
\small
\begin{split}
    \mathcal{R}_2 = \sum_{i=0}^{N-3}||(\hat{\mathbf{P}}_{t}^{i+1} &- 2\hat{\mathbf{P}}_{t}^{i} + \hat{\mathbf{P}}_{t}^{i-1})\\
    &- (\hat{\mathbf{P}}_{t-1}^{i+1} - 2\hat{\mathbf{P}}_{t-1}^{i} + \hat{\mathbf{P}}_{t-1}^{i-1})||_2.
\end{split}
\end{equation}
We observe that these regularization terms greatly improve the stability of our model outputs.

\noindent\textbf{Unsupervised Learning.}
To further improve the performance of our model when training on a dataset without point matching annotations, we employ an unsupervised approach for the correspondence learning~\cite{wang2019correspondence, wang2019unsupervisedtracking}.
By exploiting the cycle-consistency in time, we can assume that a point set tracked forward-then-backward should be matched to the point set at the same location.
To make a cycle, we run the network forward $K$ frames as usual, and then we run the network backward from the outputs of the forward pass to the initial frame. 
By doing so, we can derive matching points between the predictions during the forward and the backward pass. 
From the point set collected during the forward and backward pass~($\hat{\mathbf{P}}$ and $\hat{\mathbf{Q}}$), an unsupervised loss is defined as $\mathcal{L}_u = \frac{1}{KN} \sum_{k=0}^{K-1}\sum_{i=0}^{N-1}{||\hat{\mathbf{P}}^i_k - \hat{\mathbf{Q}}^i_k||_1}$.

\subsection{Data Augmentation by Synthetic Data}
\label{sec:synthetic_data}
Supervision for point-to-point matching is crucial for our model to learn polygonal point set tracking with accurate correspondences.
However, this information is not available in existing datasets.
To complement it, we take an image instance segmentation dataset~\cite{kuznetsova2018open_images} and synthesize video data with full supervision signals.
We first crop one or two objects from randomly sampled images in the dataset. For each cropped object, we extract a polygonal point set and deform it using the moving least squares method~\cite{schaefer2006mls} with randomly chosen control points within the point set.
These deformed objects are randomly pasted using the linear blending to a background image that is also randomly sampled. 
Then, we generate a sequence with synthetic movement by applying random affine transforms to each object and background.
This procedure allows us to generate data with full supervisory signals, and we used it to augment our training data.

%% file: body/figures/overview.tex
\begin{figure*}[t]
\begin{center}

\includegraphics[width=1\linewidth]{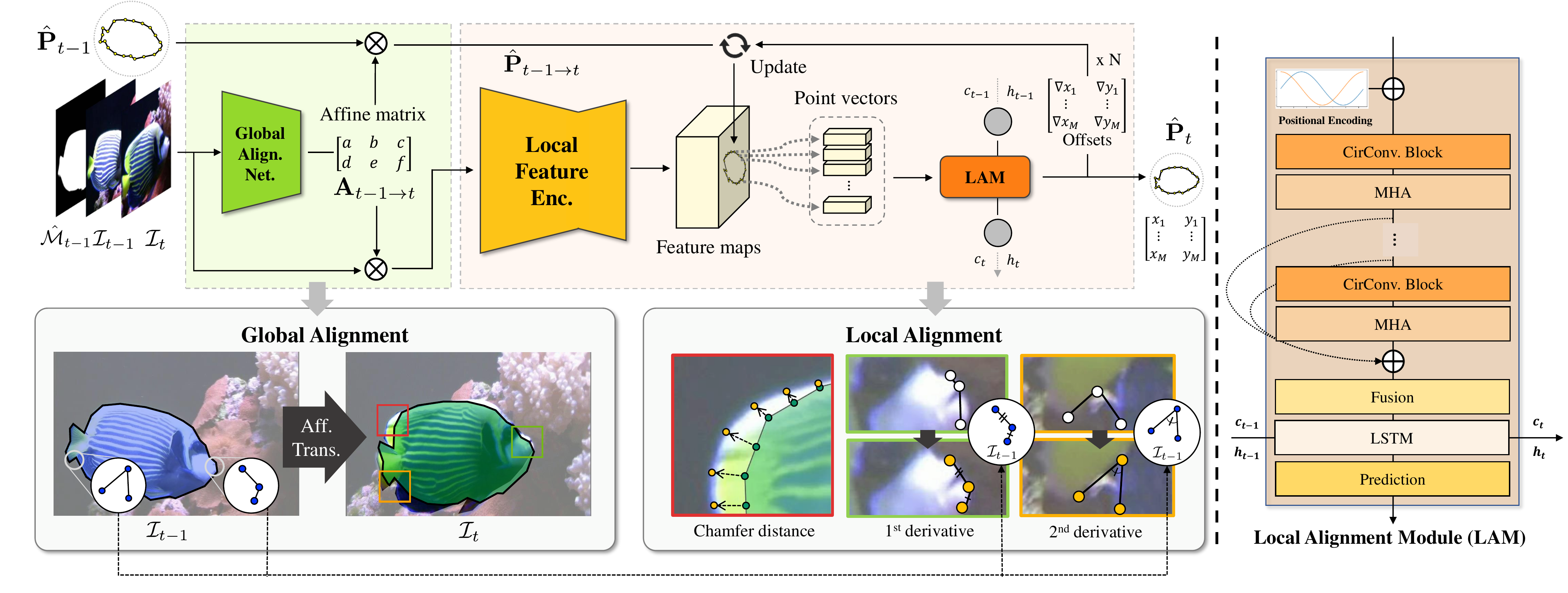}
\end{center}
\vspace*{-0.7cm}
   \caption{Overview of our framework. 
   In our framework, the alignment steps are divided into global and local alignment. 
   First, the previous frame and point set are globally aligned by the affine transform matrix from the global alignment networks.
   Then, the features encoded from the current and aligned inputs are used for local alignment after sampled as point vectors. 
   Local alignment module~(LAM) yields displacement offsets of points in the set from point vectors and updates the point set. 
   The local alignment performs iteratively in a coarse to fine manner.}
\label{fig:overview}
\vspace{-0.5cm}
\end{figure*}

%% file: body/05_experiments.tex
\section{Experiments}
\subsection{Implementation Details}
\label{sec:implementation}
To train our model, we use video object segmentation datasets, including the training sets from YouTube-VOS~\cite{xu2018youtube-vos} and DAVIS~\cite{caelles2018davis}.
We randomly choose a short clip from each sequence and extract contours from ground-truth masks using the method proposed in \cite{suzuki198findcontours}.
We filter out samples containing partial occlusion, where a contour is divided into several pieces as they cannot be tracked over frames.
From a contour, a polygonal point set is sampled for the network input.
In addition to the real data, we also make use of the synthetic dataset as described in \Sref{sec:synthetic_data}.

We train our model using Adam~\cite{kingma2014adam} optimizer with the initial learning rate of 0.0001, decayed by a factor of 10 after 70k iterations.
The backbone network of the local alignment module is initialized with pretrained weights on ImageNet~\cite{russakovsky2015imagenet}.
Our model is trained for 100k iterations with a batch size of 8 on four NVIDIA RTX 2080 Ti GPUs for 4 days. 
The number of sampled frames increases over epochs from 2 to 7.

We uniformly sample 128 vertices from the contour for the polygonal point set during both training and test time.
An input image is cropped into a patch based on the bounding box of the predicted contour with margin.
The cropped area is defined as the method proposed in \cite{bertinetto2016siamfc}.
For the local alignment, the point set is updated with 5 iterations, and we use the FPN layer with stride from 1/32 to 1/4 as our feature map for each iteration except the last iteration.
For the last iteration, the layer with stride 1/4 is used again.

\newcommand{\dataset}{PoST}
\subsection{Evaluation Dataset}

\noindent\textbf{Video Object Segmentation Datasets.}
Because our method is closely related to the video object segmentation task, we evaluate our model on DAVIS2016~\cite{caelles2018davis}, one of the most popular benchmarks for the task.
The dataset consists of 20 sequences under challenging scenarios.
Because only mask annotation is given for each target object in the dataset, we sample a point set as described in \Sref{sec:implementation}.

For the evaluation of contour-based video segmentation, Lu~\etal~\cite{lu2016cpc} proposed CPC dataset, where target object motions are mostly rigid. 
The dataset consists of 9 sequences of 34 frames in average without any training data. 
These sequences are annotated by professional designers using a standard editing tool.
The annotation is given as a parametric line of B\'{e}zier curves.

\noindent\textbf{PoST. } 
To evaluate our polygonal point set tracking properly, we need annotations not only object masks but also point correspondences across frames.
However, the existing VOS benchmark datasets aim to evaluate the quality of object masks only.
CPC~\cite{lu2016cpc} has parametric contour annotations, but its control points do not correspond with each other across frames.
Furthermore, CPC evaluation is not reliable because it only contains nine sequences, and each sequence has mostly too small motions resulting in saturated performance.

To this end, we propose a new challenging dataset for point set tracking, named as \dataset~(\textbf{Po}int \textbf{S}et \textbf{T}racking).
We take a few sequences from the existing datasets of DAVIS~\cite{caelles2018davis}, CPC~\cite{lu2016cpc}, SegTrack v2~\cite{li2013segtrack_v2} and JumpCut\cite{fan2015jumpcut} in order to cover various target object classes in different video characteristics.
To ensure that the point set tracking is possible, we avoid sequences where a target object exhibits extreme occlusions.
For each sequence, we annotate point set correspondences every 10 frames throughout the sequence. If there are no accurate corresponding points in a specific frame, we marked the points and excluded them from the evaluation. In the end, we annotated 30 sequences and use this dataset as our main benchmark.  


\subsection{Metrics}
For the evaluation on video object segmentation datasets (CPC and DAVIS2016), we use the region similarity $\mathcal{J}$ and boundary accuracy $\mathcal{F}$.
In addition, we measure \textit{average accuracy} of pixel-level mask prediction introduced in \cite{perez2019roam}.

To evaluate the point tracking, we modify the metrics, spatial accuracy~(SA) and temporal accuracy~(TA), introduced in \cite{lu2016cpc}.
These metrics measure the contour tracking accuracy by computing the distance from the ground truth points to their closest points in the predicted contour.
Different from the original metrics, we measure the distance between the exact corresponding points as follows:
\begin{equation}
\begin{split}
    \text{SA}_{\tau}(\textbf{P}, \textbf{Q}) = \lambda\sum_{t=0}^{T-1}\sum_{i=0}^{N-1}&||\textbf{P}_t^i - \textbf{Q}_t^i||_2 < \tau, \\
    \text{TA}_{\tau}(\textbf{P}, \textbf{Q}) = \lambda\sum_{t=1}^{T-1}\sum_{i=0}^{N-1}&||(\textbf{P}_t^i - \textbf{Q}_t^i) \\ 
    &- (\textbf{P}_{t-1}^i-\textbf{Q}_{t-1}^i)||_2 < \tau,
\end{split}
\end{equation}
where $\textbf{P}_t^i$ and $\textbf{Q}_t^i$ are the corresponding  points with index $i$ each in the prediction and ground-truth sets at time $t$, and $\tau$ and $\lambda$ denote a relative spatial threshold and an averaging scale factor of $\frac{1}{TN}$.


\subsection{Ablation Study}
To verify the importance of our proposals, we conduct an ablation study on three main components: local alignment module~(LAM), unsupervised loss~(UL), and synthetic data~(SD).
For a baseline, we use our local alignment network only with circular convolution blocks after the global alignment.

We perform the ablation study on \dataset~and summarize the results in \Tref{table:pointtrack}.
For better analysis, spatial accuracy~(SA) and temporal accuracy~(TA) are measured with multiple threshold settings.
Throughout all experiments, when we use the synthetic data, point tracking performance increases dramatically with an absolute gain of more than 10 points on average in terms of $\text{SA}_{.04}$. 
Under the condition of the absence of point matching loss~(without UL and SD), positional encoding and temporal information transfer in LAM does not improve performance due to no matching point supervision~(see row 1 and 3).
However, this additional information enhances point tracking performance when given point matching supervision by an absolute gain of 3 points on average in terms of $\text{SA}_{.04}$~(see row 2 and 5).
Without the point supervision, the cycle consistency of unsupervised loss only increases SA since it helps to recover the point correspondence when tracking is failed (see row 3 and 4). 
The unsupervised loss, however, further improves the performance by exploiting the synthetic data for the point matching supervision~(see row 5 and 6).

\input{body/tables/pointtrack}

\input{body/tables/pointtrack_vsothers}
\input{body/tables/cpc}

\subsection{Comparison with Other Methods}
\noindent\textbf{PoST. }
We found only few works~\cite{li2016roto++, perez2019roam} that perform a point tracking mechanism in their framework.
For completeness, we also compare our method against alternative methods such as optical flow-based tracking and patch tracking combined with a video object segmentation technique.
In the case of optical flow, the given point set is tracked through all frames by propagating each point following the flow map from a current state-of-the-art optical flow method~\cite{zhao2020maskflownet}. 
Patch tracking can also be used to track each point by extracting patches centered on given points.
For the patch tracking method, we can additionally guide the result to stick to the object boundary using object masks from the-state-of-the-art video object segmentation method, STM~\cite{oh2019stm}.
We use CSRT~\cite{lukezic2017csrt} for the patch tracking here.

Results of each method on PoST is reported in \Tref{table:pointtrack_vsothers}.
As the results show, our model yields superior performance compared with existing competitors in point set tracking in terms of SA and TA with entire thresholds.
Note that patch tracking shows impressive results in terms of SA, but the outlines of the object were not preserved well in this setting.

\input{body/tables/davis}
\noindent\textbf{Video Object Segmentation Datasets. }
Although video object segmentation is not the main goal in this paper, our method can represent object contour using a polygonal point set.
For evaluation, we report performance on video object segmentation dataset, CPC~\cite{lu2016cpc} and DAVIS2016~\cite{caelles2018davis}.

Quantitative results on CPC are summarized in \Tref{table:cpc}.
In CPC, even though the performance is saturated due to the rigidity and the static motion of the targets in the dataset, our method shows a comparable accuracy with other methods. 
Note that STM~\cite{oh2019stm} aims to derive a region-based object mask thus does not track a point set.
We report the performance of CPC~\cite{lu2016cpc} only for a subset of the dataset as reported in the original paper, because their codes are not available.
In Roto++, we also test with two ground-truth keyframes of the first and last frames because interpolation between keyframes plays an important role in the mechanism.
The performance of ROAM~\cite{perez2019roam} is reported in two cases: the number reported in the original paper and our reproduced results with an official code in the default setting.

For DAVIS2016 validation set, the performance of various methods is shown in \Tref{table:davis}.
Since the dataset targets non-rigid object motion and occlusions, contour-based object segmentation methods inherently have many challenges on this dataset.
Despite the limitation, our method outperforms other point set tracking approaches with significant gaps in all metrics. 

\noindent\textbf{Runtime Performance.}
\Tref{table:runtime} shows runtime performance on DAVIS2016. 
We use GeForce TITAN X as in \cite{perez2019roam}.
Our method is the fastest one among other point set tracking approaches.

\input{body/figures/qualitative}
\input{body/figures/applications}

\subsection{Qualitative Results}
\Fref{fig:qualitative} shows some qualitative results of our method.
Given an initial polygonal point set of the target object, our model propagates the set over frames.
Each corresponding point is marked with unique indices in different colors. 
Our model yields successful results in terms of both regional segmentation and point set propagation for the target object through all sequences.

\input{body/tables/runtime}
\subsection{Applications}
\Fref{fig:application} showcases some applications of the point set tracking.
We can apply text mapping by motion tracking for a rigid object in \Fref{fig:application}~(a) and part-level deformation for non-rigid object in \Fref{fig:application}~(b).
Our method makes it easy to apply these effects, whereas patch tracking and region-based segmentation require additional information.

%% file: body/tables/pointtrack.tex
\begin{table}[t]

\begin{center}
\small
\setlength{\tabcolsep}{5pt}
\begin{adjustbox}{width=1\linewidth}

\begin{tabular}{ccc|ccc|ccc}
\toprule
LAM & UL & SD   & SA$_{.16}$    & SA$_{.08}$    & SA$_{.04}$  & TA$_{.16}$    & TA$_{.08}$    & TA$_{.04}$  \\
\midrule

            &                     &              & 0.906 & 0.776 & 0.615 & \underline{0.976} & \underline{0.943} & 0.846  \\
            &                     & \checkmark             & 0.907 & 0.820 & 0.701 & 0.974 & \underline{0.943} & 0.881  \\
\checkmark            &                     &             & 0.909 & 0.774 & 0.599 & 0.969 & 0.924 & 0.819  \\
\checkmark            & \checkmark                    &             & 0.909 & 0.808 & 0.672 & 0.961 & 0.920 & 0.829  \\
\checkmark            &                     & \checkmark             & \underline{0.950} & \underline{0.865} & \underline{0.736} & \textbf{0.977} & \underline{0.943} & \underline{0.884}  \\

\midrule

\checkmark            & \checkmark                    & \checkmark               & \textbf{0.964} & \textbf{0.902} & \textbf{0.803} & \textbf{0.977} & \textbf{0.956} & \textbf{0.896}  \\
\bottomrule
\end{tabular}
\end{adjustbox}
\end{center}
\vspace{-0.5cm}
\caption{Ablation studies on \dataset. Three different components of our framework are validated: local alignment module~(LAM), unsupervised learning by temporal cycle consistency~(UL) and synthetic data~(SD).}
\label{table:pointtrack}
\vspace{-0.2cm}
\end{table}

%% file: body/tables/pointtrack_vsothers.tex
\begin{table}[t]

\begin{center}
\setlength{\tabcolsep}{3pt}
\begin{adjustbox}{width=1\linewidth}
\small
\begin{tabular}{l|ccc|ccc}
\toprule
Method & SA$_{.16}$    & SA$_{.08}$    & SA$_{.04}$     & TA$_{.16}$    & TA$_{.08}$    & TA$_{.04}$ \\
\midrule
CSRT~\cite{lukezic2017csrt} & \underline{0.925} & \underline{0.878} & \textbf{0.807} & \underline{0.973} & \underline{0.927} & \underline{0.842} \\
\midrule
MaskFlownet~\cite{zhao2020maskflownet} & 0.664 & 0.497 & 0.345 & 0.941 & 0.831 & 0.640 \\


STM~\cite{oh2019stm} + CSRT~\cite{lukezic2017csrt} & 0.856 & 0.715 & 0.550 & 0.965 & 0.910 & 0.815 \\ 

Roto++~\cite{li2016roto++} & 0.769 & 0.530 & 0.366 & 0.841 & 0.630 & 0.403 \\

ROAM~\cite{perez2019roam}& 0.871 & 0.717 & 0.512 & 0.965 & 0.873 & 0.697 \\ 
\midrule

\textbf{Ours}              & \textbf{0.964} & \textbf{0.902} & \underline{0.803} & \textbf{0.977} & \textbf{0.956} & \textbf{0.896} \\

\bottomrule
\end{tabular}
\end{adjustbox}
\end{center}
\vspace{-0.5cm}
\caption{Comparison with other methods on \dataset.}
\label{table:pointtrack_vsothers}
\vspace{-0.2cm}
\end{table}

%% file: body/tables/cpc.tex
\begin{table}[t]

\begin{center}
\setlength{\tabcolsep}{10pt}
\small
\begin{tabular}{l|ccc}

\toprule
& \makecell{Avg.\\Acc.} & \makecell{$\mathcal{J}$} & \makecell{$\mathcal{F}$}\\
\midrule
STM~\cite{oh2019stm} & \underline{0.997} & \underline{0.957} & 0.982 \\
\midrule
MaskFlownet~\cite{zhao2020maskflownet} & 0.948 & 0.625 & 0.627 \\
CPC$^*$~\cite{lu2016cpc} & \textbf{0.998}$^*$ & \textbf{0.963}$^*$ & \textbf{0.997}$^*$ \\
Roto++~(1 kf)~\cite{li2016roto++} & 0.976 & 0.640 & 0.527 \\
Roto++~(2 kf)~\cite{li2016roto++} & 0.989 & 0.840 & 0.810\\
ROAM~\cite{perez2019roam} & 0.995 & 0.951 & - \\
ROAM$^\dag$~\cite{perez2019roam} & 0.995$^\dag$ & 0.859$^\dag$ & 0.893$^\dag$ \\
\midrule
\textbf{Ours} & \underline{0.997} & 0.948 & \underline{0.995} \\
\bottomrule
\end{tabular}
\end{center}
\vspace{-0.5cm}
\begin{flushright}\footnotesize{$^*$ partial evaluation.}\end{flushright}
\vspace{-0.7cm}
\begin{flushright}\footnotesize{$^\dag$ reproduced with default setting.}\end{flushright}
\vspace{-0.5cm}
\caption{Quantitative Results on CPC.}
\label{table:cpc}
\vspace{-0.5cm}
\end{table}

%% file: body/tables/davis.tex
\begin{table}[t]

\begin{center}
\setlength{\tabcolsep}{10pt}


\small
\begin{tabular}{l|ccc}

\toprule
& \makecell{Avg.\\Acc.} & \makecell{$\mathcal{J}$} & \makecell{$\mathcal{F}$}\\
\midrule
STM~\cite{oh2019stm} & \textbf{0.992} & \textbf{0.887} & \textbf{0.899} \\
\midrule
MaskFlownet~\cite{zhao2020maskflownet} & 0.873 & 0.300 & 0.289 \\
Roto++~(1 kf)~\cite{li2016roto++} & 0.908 & 0.217 & 0.195 \\
Roto++~(2 kf)~\cite{li2016roto++} & 0.926 & 0.329 & 0.290 \\
ROAM~\cite{perez2019roam} & 0.952 & 0.583 & - \\
ROAM$^\dag$~\cite{perez2019roam} & 0.929$^\dag$ & 0.378$^\dag$ & 0.335$^\dag$ \\
\midrule
\textbf{Ours} & \underline{0.971} & \underline{0.642} & \underline{0.637} \\
\bottomrule
\end{tabular}

\end{center}
\vspace{-0.5cm}
\begin{flushright}\footnotesize{$^\dag$ reproduced with default setting.}\end{flushright}
\vspace{-0.6cm}
\caption{Quantitative Results on DAVIS2016 Val.}
\label{table:davis}
\vspace{-0.6cm}
\end{table}

%% file: body/figures/qualitative.tex
\begin{figure*}[t]
\begin{center}

\includegraphics[width=1\linewidth]{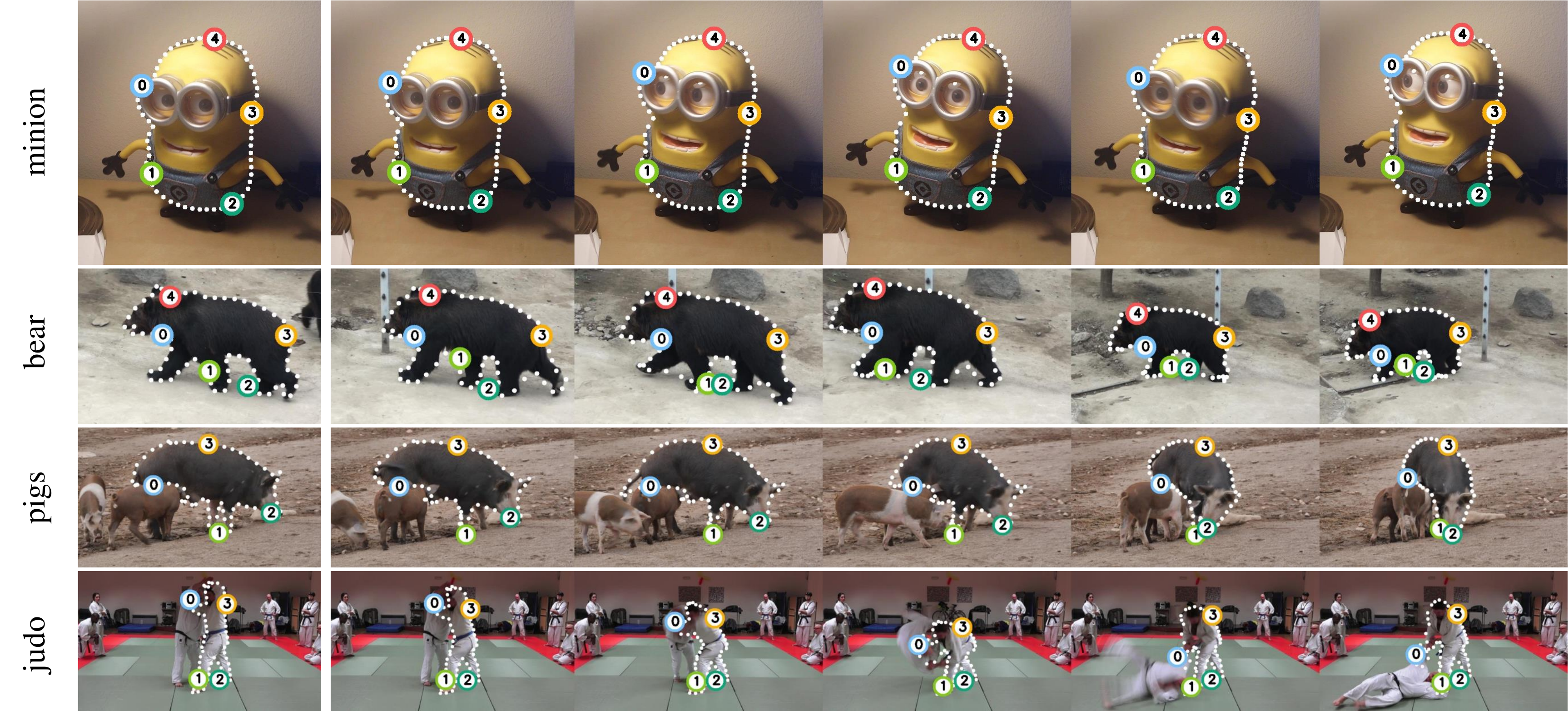}
\end{center}
\vspace*{-0.6cm}
   \caption{Qualitative results on various datasets. The images in the first column are the initial frames of each clip.
   Points in a predicted point set are colored as white and several identical points with the same index are visualized in the same color. 
   For better analysis, we select samples in different categories and scenarios.}
\vspace*{-0.3cm}
\label{fig:qualitative}
\end{figure*}

%% file: body/figures/applications.tex
\begin{figure*}[t]
\begin{center}

\includegraphics[width=1\linewidth]{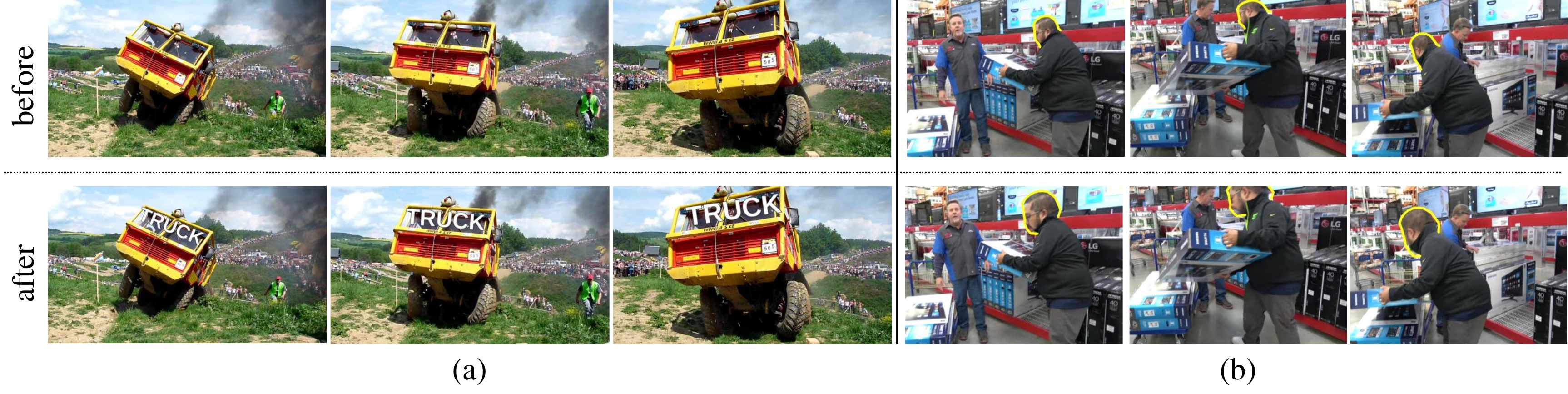}
\end{center}
\vspace*{-0.7cm}
   \caption{Applications of our point set tracking method. (a) Text is mapped by motion tracking in front of the truck. (b) The head of a man is exaggerated by part distortion.}
\vspace*{-0.5cm}
\label{fig:application}
\end{figure*}

%% file: body/tables/runtime.tex
\begin{table}[t]
\begin{center}

\setlength{\tabcolsep}{3pt}
\begin{adjustbox}{width=1\linewidth}
\small
\begin{tabular}{l|cccc}
\toprule
& Roto++~\cite{li2016roto++} & ROAM~\cite{perez2019roam} & STM+CSRT~\cite{oh2019stm,lukezic2017csrt} & \textbf{Ours} \\ 
\midrule
Time (ms) / Frame & 118$^*$ & 5639$^*$ & 2158 & 84 \\
\bottomrule
\end{tabular}
\end{adjustbox}
\end{center}
\vspace{-0.5cm}
\begin{flushright}\footnotesize{$*$ reported in \cite{perez2019roam}.}\end{flushright}
\vspace{-0.5cm}
\caption{Runtime Performances on DAVIS2016 Val.}
\label{table:runtime}
\vspace{-0.6cm}
\end{table}

%% file: body/06_conclusion.tex
\section{Conclusion}

In this paper, we proposed a learning-based method for polygonal point set tracking.
We designed global and local alignment networks for polygonal point set tracking. To train the network, we introduced our learning scheme using synthetic data and the unsupervised cycle-consistency loss. We demonstrated that our model successfully propagates a polygonal point set over time with accurate point-wise correspondences even without any fully annotated ground-truth data. 
We achieved state-of-the-art performance on multiple benchmarks and showcased interesting applications.

%% file: body/07_acknowledgements.tex
\noindent\textbf{Acknowledgements} This work was supported in part by the Institute of Information and Communications Technology Planning and Evaluation~(IITP) Grant funded by the Korean Government~(MSIT), Artificial Intelligence Graduate School Program, Yonsei University, under Grant 2020-0-01361.

%% file: supple/supple_main.tex
\begin{center}
  \Large\textbf{Supplementary Materials: \\ Polygonal Point Set Tracking}
\end{center}
\vspace*{\stretch{0.5}}

This supplementary material provides details of the proposed evaluation dataset, named as PoST, synthetic data, and more quantitative and qualitative results. 
In \Sref{sec:post}, we describe additional statistic details of PoST and visualize sample sequences.
We introduced the method to synthesize data for data augmentation in the main paper.
In \Sref{sec:synthetic}, some samples of the synthetic data are provided.
In \Sref{sec:quantitative} and \Sref{sec:qualitative}, we provide more detailed quantitative and qualitative results and compare with other methods.

\section{Details of PoST}
\input{body/tables/statistics}
\input{body/figures/pointtrack}
\label{sec:post}
For better analysis, we report statistical details of PoST in \Tref{table:stat}.
The table provides six properties for each sequence: the number of ground-truth points, input size, sequence length, motion~(MO), scale change~(SC) and occluded points~(OC).

We measure camera and object motion~(MO) by computing the distance between corresponding points as follows:
\begin{equation}
\label{eq:motion}
    \text{MO} = \frac{1}{TN}\sum_{t=1}^{T-1}\sum_{i=0}^{N-1}{||\mathbf{P}_{t}^{i} - \mathbf{P}_{t-1}^{i} ||_2},
\end{equation}
where $\mathbf{P}_{t}^{i}$ denotes the $i^{th}$~($0 \leq i < N$) normalized ground-truth point of the $t^{th}$~($0 \leq t < T$) frame and $||\cdot||_2$ refers to the Euclidean distance between points.
Scale change~(SC) is one of the challenging scenarios in the point tracking. 
We measure the changes by computing temporal changes of distance between adjacent points as follows:
\begin{equation}
    \text{SC} = \frac{1}{TN}\sum_{t=1}^{T-1}\sum_{i=0}^{N-1}{\bigg|||\mathbf{P}_{t}^{i} - \mathbf{P}_{t}^{i-1}||_2 - ||\mathbf{P}_{t-1}^{i} - \mathbf{P}_{t-1}^{i-1}||_2\bigg|}.
\end{equation}
Since we skip annotating the ambiguous corresponding points due to self-occlusion, we also report the average number of these points in each sequence~(OC).


Some examples of PoST are shown in \Fref{fig:pointtrack}.
We sample seven random points and each corresponding point is represented by a unique index of a different color.
The green outline in the first frame indicates the ground-truth contour of the target object provided in the dataset.
The missing points in some frames imply the ambiguous points due to self-occlusion.

\section{Samples of Synthesizing Data}
\label{sec:synthetic}
\input{body/figures/synthetic}
Data augmentation by synthesizing image segmentation data from \cite{kuznetsova2018open_images} is one of the important features to enhance the point tracking performance in our framework.
For clarification of the synthetic data, some samples are shown in \Fref{fig:synthetic}.
Note that the black areas are the result of random transformation to simulate the movement of the scene and the objects.
Ground-truth corresponding point samples are marked in the same way as done in \Fref{sec:post}, and the white-colored dense points are also added.
Supervision for the point correspondence is provided by the synthetic data.
Since we transform the original point set along with the image, the point correspondence is maintained over the entire frame.

\section{Quantitative Result Analysis}
\label{sec:quantitative}
\input{body/figures/plots}
In \Fref{fig:plots}, we evaluate our method and the other competitors~\cite{li2016roto++, perez2019roam, lukezic2017csrt, oh2019stm} using spatial and temporal accuracy~(SA, TA) with various thresholds.
Following Equation~(9) in the main paper, the spatial accuracy directly measures the distance between the ground-truth and the predicted point while the temporal accuracy implies the shape tracking performance.
For both metrics, our method achieves the best performance compared to the other competitors for all thresholds. 
Specifically, for the spatial accuracy, our method outperforms the other methods with significant margins even using a high threshold.
Also, the difference in saturation tendency of the two metrics indicates that our approach exceeds the other methods, especially in terms of the point correspondence.


\section{Qualitative Result Comparison}
\label{sec:qualitative}
\input{body/figures/supp_comparison}
\Fref{fig:supp_comparison} provides some qualitative results of our method and the other methods for comparison.
We sample the sequences from the datasets of PoST~(pot), DAVIS2016~\cite{caelles2018davis}~(blackswan, camel) and CPC~\cite{lu2016cpc}~(plane).
Visualization of the results is done in the same way as \Fref{fig:synthetic}.
While Roto++~\cite{li2016roto++} and ROAM~\cite{perez2019roam} suffer drift problem and show inferior performance in terms of point correspondence, our method estimates the point set location more stably.
Note that we use the first and the last frames as the key frame for Roto++~(2kf).

\section{Cross-domain Evaluation}
\label{sec:cross-domain}
\input{body/figures/cross-doamin}
Because our method can track points containing low-level features, robustness in cross-domain prediction is expected.
To prove this hypothesis, we test our model trained as described in the main paper on cell tracking sequences published in \cite{ulman2017cell_tracking}. 
Cell tracking images belong to a different domain from the training data, in that the images are grayscale and the edge of the target deforms unstably.
As \Fref{fig:cross-domain} shows, while a state-of-the-art video object segmentation method~\cite{oh2019stm} fails to predict the desired object, our method tracks cells successfully.
We only report the early four frames for \cite{oh2019stm} since the mask prediction is failed following frames.
For our method, we visualize the results on the following frames separately in the third partition.

%% file: body/tables/statistics.tex
\begin{table*}[t]

\begin{center}
\setlength{\tabcolsep}{9pt}

\begin{subtable}{1.0\textwidth}
\begin{adjustbox}{width=1\linewidth}
\begin{tabular}{l|cccccccccc}
\toprule
{} &     bear &  blackswan &      boy & car-roundabout & car-shadow &   cheetah &      cup &      drop &     fish &  freeway \\
\midrule
\# of points &       10 &         18 &       18 &             15 &         12 &        21 &       12 &         6 &       10 &        6 \\
Input size  &  640x400 &  1920x1080 &  384x540 &      1920x1080 &  1920x1080 &  1280x720 &  600x534 &  1280x718 &  720x480 &  762x506 \\
Length      &      151 &         41 &       21 &             71 &         31 &       181 &      371 &        21 &       81 &       31 \\
MO          &    0.049 &      0.035 &    0.037 &          0.084 &      0.051 &     0.048 &    0.068 &     0.138 &    0.060 &    0.049 \\
SC          &    0.014 &      0.003 &    0.011 &          0.011 &      0.014 &     0.008 &    0.006 &     0.005 &    0.014 &    0.006 \\
OC          &    0.000 &      0.000 &    0.000 &          0.875 &      0.000 &     1.368 &    0.684 &     0.000 &    0.000 &    0.000 \\
\bottomrule
\end{tabular}
\end{adjustbox}
\end{subtable}
\medskip

\begin{subtable}{1.0\textwidth}
\begin{adjustbox}{width=1\linewidth}
\begin{tabular}{l|cccccccccc}
\toprule
{} &  giraffe & helicopter &   hiphop &    labcoat &   minion &   monkey & monkey-head & monkey-horse &  penguin &      pig \\
\midrule
\# of points &       11 &         17 &       11 &         16 &       17 &       10 &          13 &           20 &       11 &        8 \\
Input size  &  658x484 &  1920x1080 &  960x540 &  3840x2160 &  540x575 &  480x270 &     960x540 &      960x540 &  384x212 &  695x480 \\
Length      &      191 &         41 &       81 &         41 &       31 &       31 &          31 &           31 &       41 &      281 \\
MO          &    0.026 &      0.157 &    0.159 &      0.249 &    0.029 &    0.060 &       0.026 &        0.039 &    0.021 &    0.114 \\
SC          &    0.005 &      0.011 &    0.018 &      0.093 &    0.006 &    0.023 &       0.003 &        0.004 &    0.007 &    0.011 \\
OC          &    0.100 &      1.000 &    0.778 &      4.600 &    0.000 &    0.000 &       0.500 &        0.000 &    0.000 &    0.517 \\
\bottomrule
\end{tabular}
\end{adjustbox}
\end{subtable}
\medskip

\begin{subtable}{1.0\textwidth}
\begin{adjustbox}{width=1\linewidth}
\begin{tabular}{l|cccccccccc}
\toprule
{} &     plane &      pot &   skater & slackline &  soldier &  station &   sunset &     tower &      toy &     worm \\
\midrule
\# of points &        35 &        7 &       35 &        16 &       23 &       16 &        3 &        22 &       16 &       11 \\
Input size  &  1488x914 &  960x540 &  720x480 &  1920x816 &  528x224 &  656x492 &  960x540 &  1280x720 &  960x540 &  480x264 \\
Length      &        31 &      241 &       81 &        51 &       31 &      371 &       31 &        21 &      351 &       81 \\
MO          &     0.036 &    0.037 &    0.235 &     0.567 &    0.339 &    0.238 &    0.017 &     0.002 &    0.029 &    0.045 \\
SC          &     0.006 &    0.007 &    0.012 &     0.033 &    0.019 &    0.007 &    0.029 &     0.001 &    0.003 &    0.012 \\
OC          &     0.000 &    0.360 &   11.111 &     5.667 &    2.500 &    2.526 &    0.000 &     0.000 &    0.167 &    0.000 \\
\bottomrule
\end{tabular}
\end{adjustbox}
\end{subtable}

\end{center}
\vspace{-0.5cm}
\caption{The statistical details of PoST. We report six properties for each sequence: the number of ground-truth points, input size, sequence length~(\ie the number of frames), motion~(MO) including both of camera and object motion, scale change~(SC) and self-occluded points per frame~(OC).}
\label{table:stat}
\end{table*}

%% file: body/figures/pointtrack.tex
\begin{figure*}[t]
\begin{center}

\includegraphics[width=1.0\linewidth]{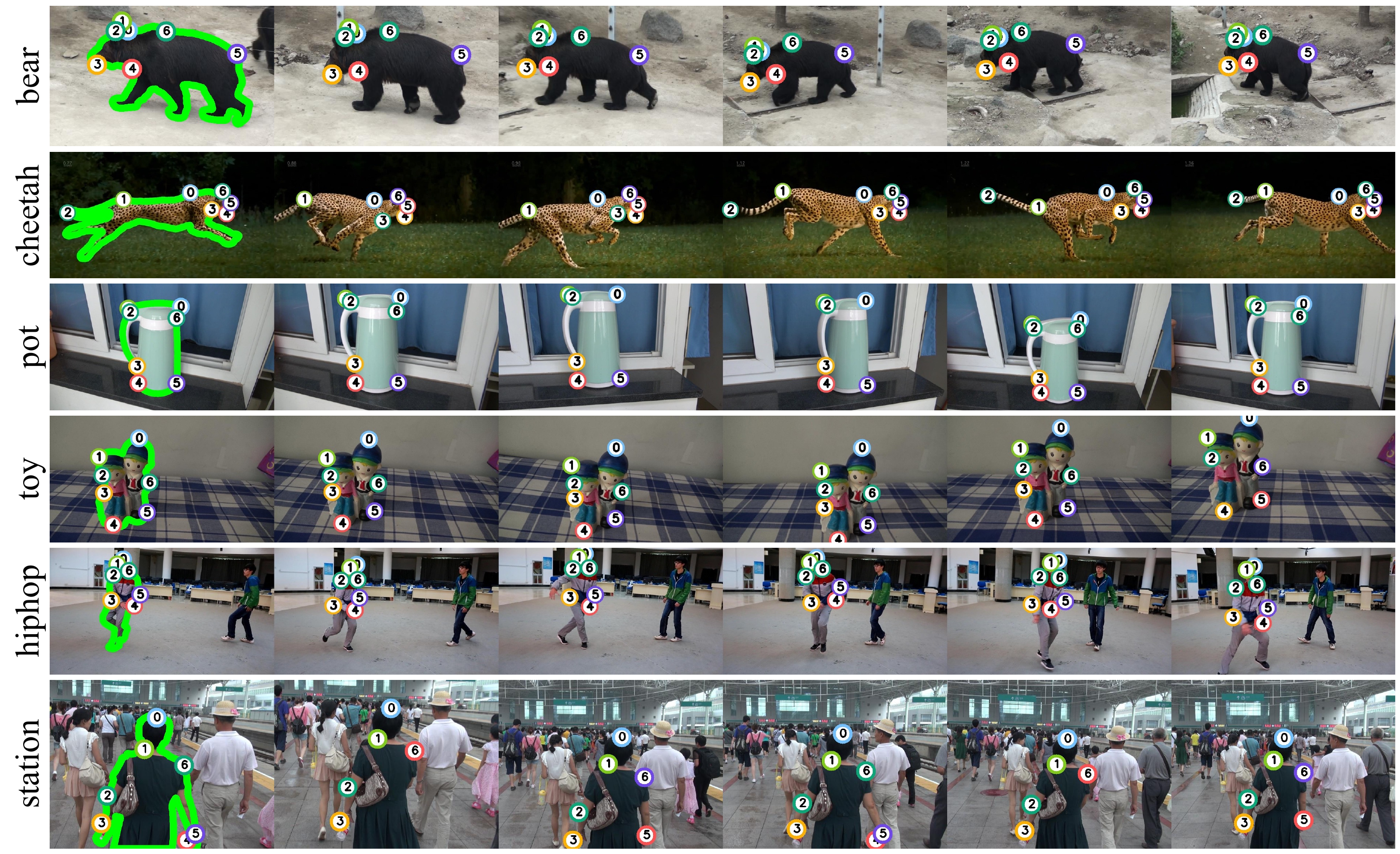}
\end{center}
\vspace{-0.5cm}
   \caption{Examples of PoST evaluation dataset. Randomly sampled seven points are marked with the same colored corresponding points as the same indices.}
\vspace{-0.5cm}
\label{fig:pointtrack}
\end{figure*}

%% file: body/figures/synthetic.tex
\begin{figure}[t]
\begin{center}

\includegraphics[width=1.0\linewidth]{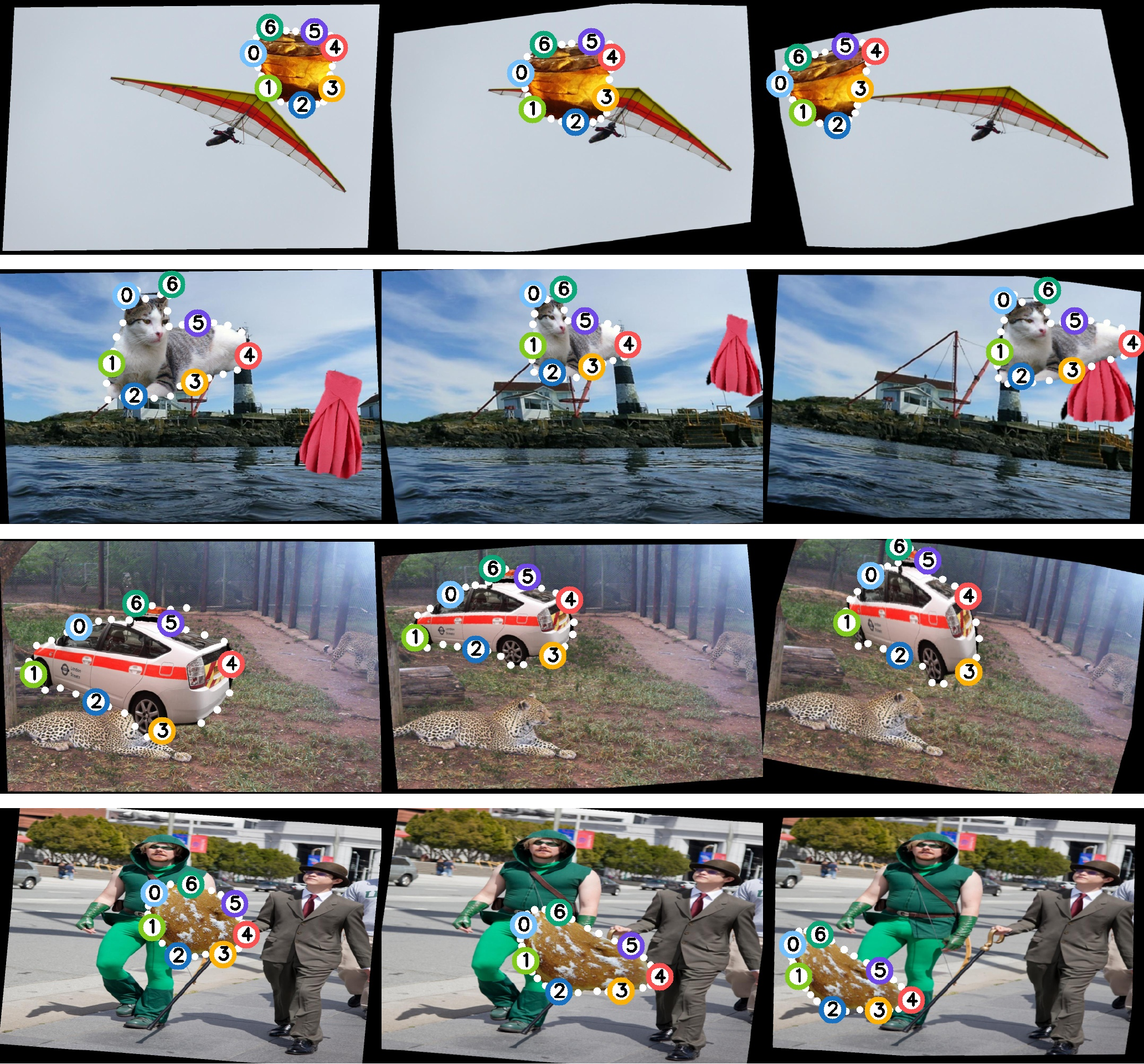}
\end{center}
   \caption{Example of synthetic dataset.}
\label{fig:synthetic}
\end{figure}

%% file: body/figures/plots.tex
\begin{figure}[h]
\begin{center}
\begin{subfigure}{1.0\linewidth}
\vspace{-0.2cm}
\includegraphics[width=1.0\linewidth]{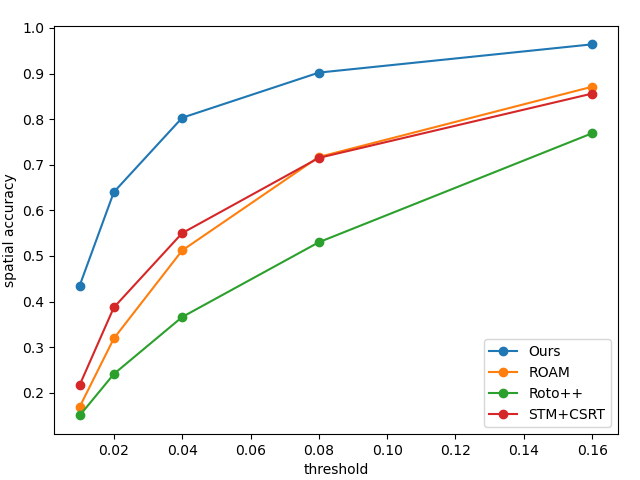}
   \caption{Spatial Accuracy~(SA).}
\label{fig:plot_sa}
\end{subfigure}
\begin{subfigure}{1.0\linewidth}
\includegraphics[width=1.0\linewidth]{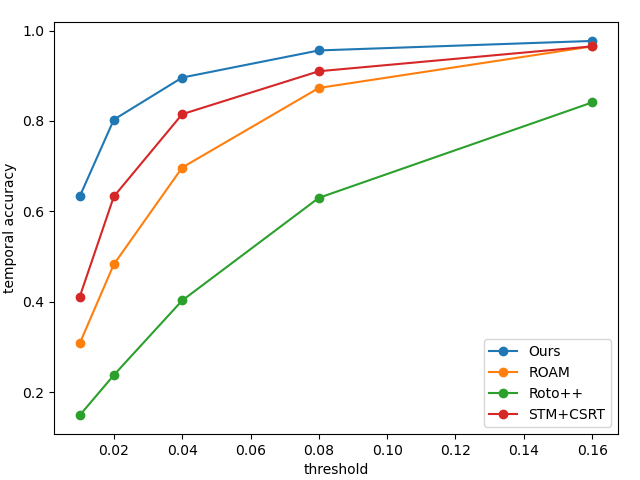}
   \caption{Temporal Accuracy~(TA).}
\vspace{-0.1cm}
\label{fig:plot_ta}
\end{subfigure}
\end{center}

\caption{Plots of each metrics on different thresholds. Our method outperforms other competitors over entire thresholds.}
\vspace{-0.5cm}
\label{fig:plots}
\end{figure}

%% file: body/figures/supp_comparison.tex
\begin{figure*}[t]
\begin{center}

\includegraphics[width=1.0\linewidth]{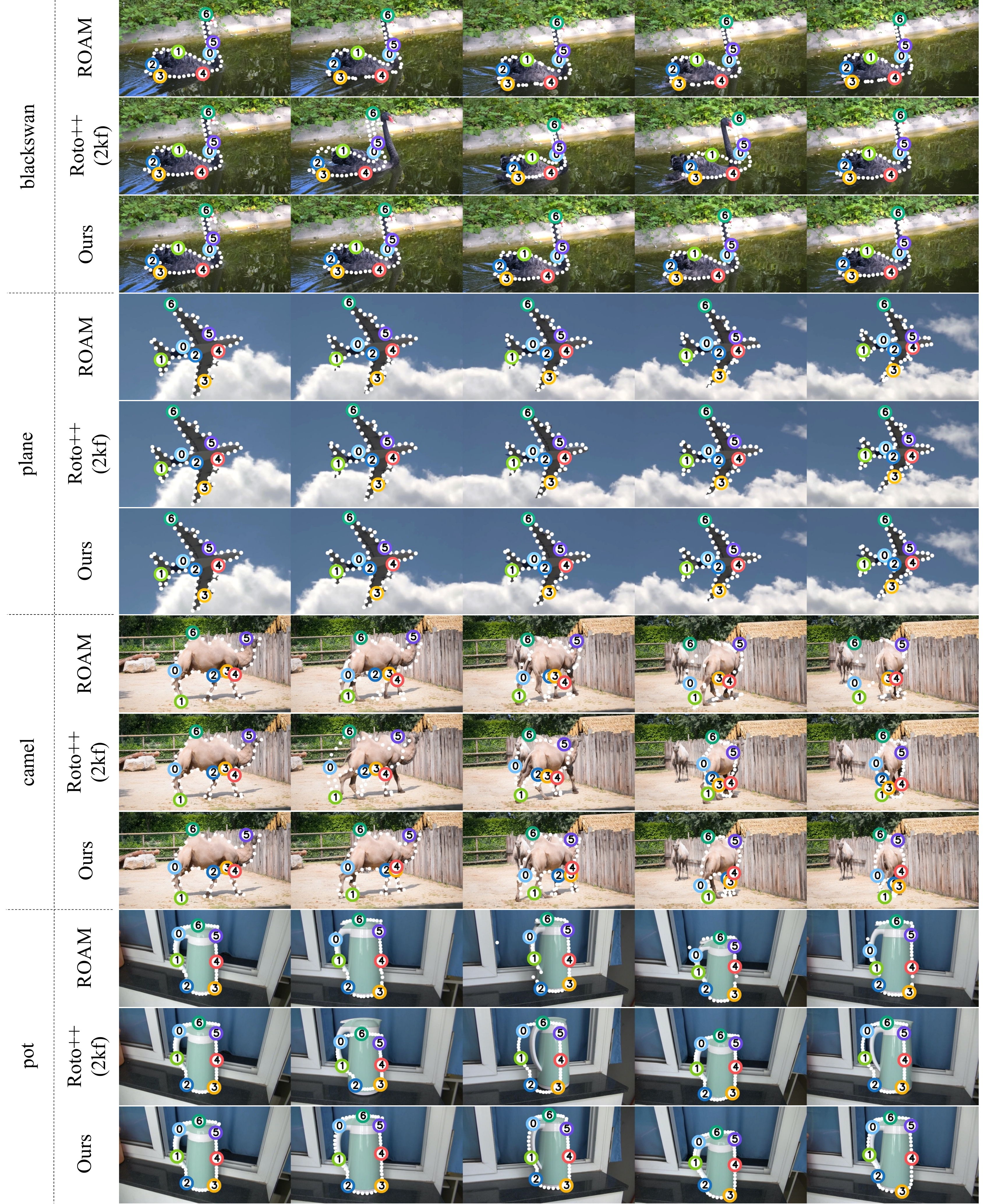}
\end{center}
\vspace{-0.5cm}
   \caption{Qualitative results of our method, Roto++ and ROAM on various datasets. While the other methods do not take the point correspondence into account, our method estimates the correspondence more precisely than the others.}
\vspace{-0.5cm}
\label{fig:supp_comparison}
\end{figure*}

%% file: body/figures/cross-doamin.tex
\begin{figure}[t]
\begin{center}

\includegraphics[width=1.0\linewidth]{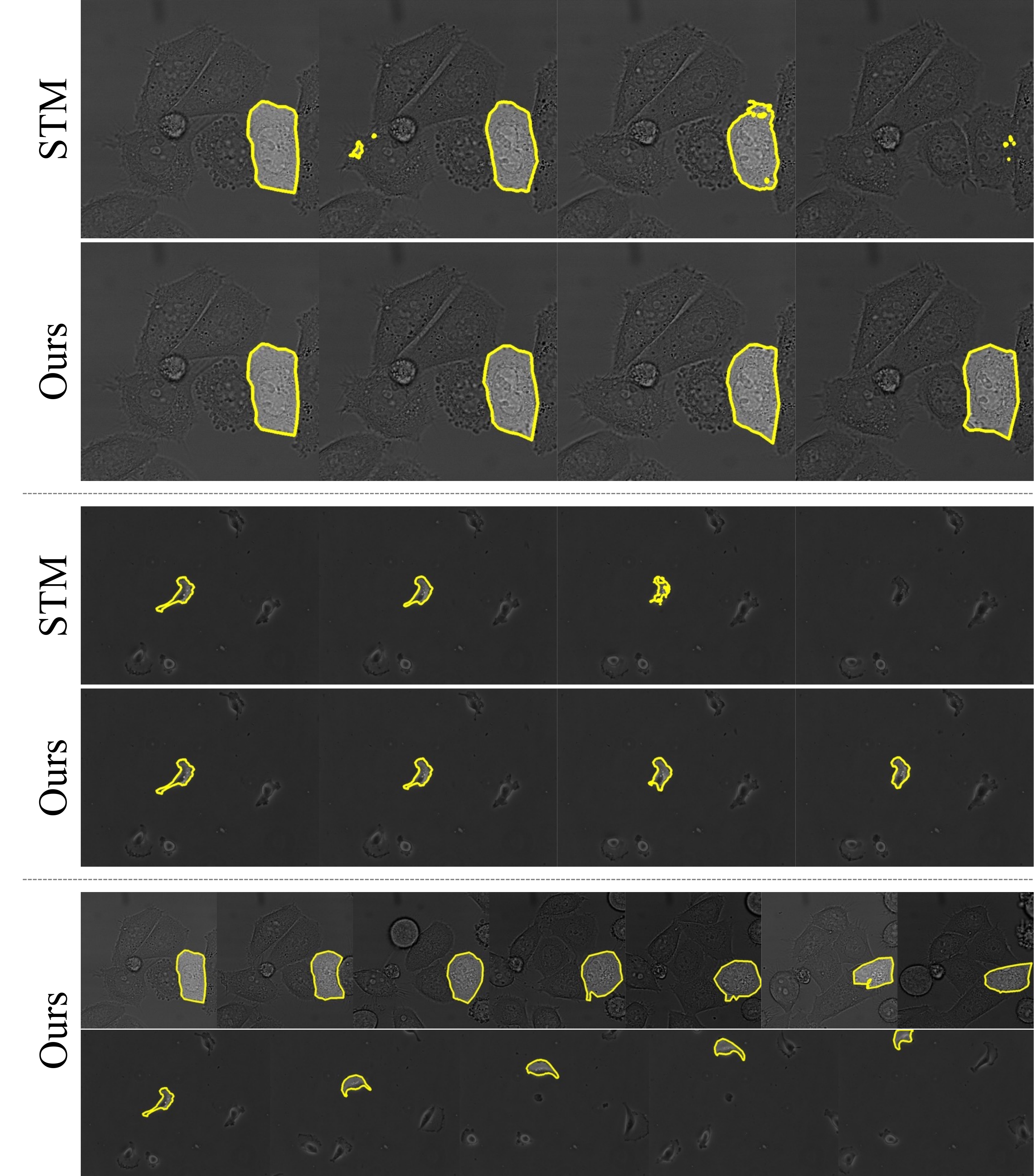}
\end{center}
\vspace{-0.2cm}
   \caption{Cell tracking for cross-domain evaluation. Comparison with a state-of-the-art method of video object segmentation is shown in the first two partitions separated by the dash lines. We visualize our results over following frames in the last partition.}
\label{fig:cross-domain}
\end{figure}